\documentclass{article}
 \PassOptionsToPackage{numbers, compress}{natbib}
\usepackage[final]{neurips_2020}

\usepackage[utf8]{inputenc} 
\usepackage[T1]{fontenc}    
\usepackage{hyperref}       
\usepackage{url}            
\usepackage{booktabs}       
\usepackage{amsfonts}       
\usepackage{nicefrac}       
\usepackage{microtype}      
\usepackage{color}
\usepackage[dvipsnames]{xcolor}
\usepackage{amsmath}
\usepackage{amssymb}
\usepackage{pifont}
\usepackage{graphicx}
\usepackage{subcaption}
\usepackage{multirow}
\usepackage[normalem]{ulem}
\usepackage{enumitem}
\usepackage{wrapfig}

\usepackage{hyperref}

\title{
Balance Regularized Neural Network Models for Causal Effect Estimation
}

\author{%
Mehrdad Farajtabar\thanks{Correspondence to 
\href{mailto:farajtabar@google.com}{farajtabar@google.com} }
  \And
  Andrew Lee \\
  \And
  Yuanjian Feng \AND
  \And
  Vishal Gupta  \\ 
  \And
  Peter Dolan \\
  \And
  Harish Chandran \\
  \And
  Martin Szummer \hspace{6mm}  \AND
  \texttt{DeepMind}
  }

\begin{document}

\maketitle

\begin{abstract}
Estimating individual and average treatment effects from observational data is an important problem in many domains such as healthcare and e-commerce. In this paper, we advocate balance regularization of multi-head neural network architectures. Our work is motivated by representation learning techniques to reduce differences  between treated and untreated distributions that potentially arise due to confounding factors.  We further regularize the model by encouraging it to predict control outcomes for individuals in the treatment group that are similar to control outcomes in the control group. We empirically study the bias-variance trade-off between different weightings of the regularizers, as well as between inductive and transductive inference.
\end{abstract}

\section{Introduction}
Causal inference practitioners are increasingly applying deep learning to estimate causal effects from observational data, capitalizing on the expressivity and representational power of neural networks.  Authors in~\citep{johansson2016learning,shalit2017estimating,shi2019adapting} propose neural network architectures with separate outputs ("heads") for each potential outcome~\cite{rubin2005causal}.  However, when learning from observational data, rather than from randomized trials, we must address potential bias due to confounding factors.  Fortunately,   neural networks allow for flexible forms of regularization. We devise novel regularizers to tackle confounding, and also leverage existing regularizers from domain adaptation and covariate shift~\cite{johansson2016learning}.

In this paper we extend recent deep causal inference methods by combining their approaches into one architecture, and study how different components affect causal effect estimation. We consider a multi-head model inspired by~\citep{shi2019adapting} that predicts the potential outcomes of data containing control and treated sets of data.
Neural network models have shown improvements in estimation accuracy, but there is little understanding of how. We elucidate them by decomposing their mean squared error improvements into bias and variance. Our contributions can be summarized in three parts. 

First, we evaluate the ability of the multi-head model to improve estimation accuracy by comparing it with a single-head alternative, and by analyzing the bias and variance of inductive and transductive formulations.

Second, we evaluate the ability of a maximum mean discrepancy loss (MMD)~\citep{gretton2012kernel} on an intermediate embedding layer to regularize a shared representation of treatment and control units  by penalizing covariate imbalance, inspired by~\cite{shalit2017estimating}. We interpret this as a form of balance regularization as the prediction task of both heads influence the balancing of covariates between treated and control groups.

Third, we introduce balance regularization with a prognostic score (PRG)~\citep{hansen08prognostic}. A prognostic score measures the similarity between the predicted untreated outcomes of treated and control groups by using standardized mean difference or a two sample Kolmogorov-Smirnov (KS) statistic, both of which can be minimized with an additional loss.

\section{Preliminaries}
The observed data $D$ consists of $n$ triplets $\{(x_i, t_i, y_i)\}_{i=1...n}$, where $x_i$ is the feature or context of the $i$-th unit, $t_i$ is the treatment (intervention), and $y_i$ is the associated outcome. For simplicity we focus on binary treatment where $t_i=\{0,1\}$. 
The potential outcome for unit $i$ is $Y_i^0$ (associated to $t=0$) or $Y_i^1$ (associated to $t=1$). The observed outcome can thus be expressed as:
\begin{equation}
y_i = t_i Y_i^1 + (1-t_i) Y_i^0.
\end{equation}
For a unit $i$ described by $x_i$ we are interested in estimating the impact of the treatment $im(x_i)$ (or individual treatment effect (ITE)), i.e., 
\begin{equation}
im(x_i) = Y_i^1(x_i) - Y_i^0(x_i).
\end{equation}
The fundamental problem in causal inference is that only one of the two potential outcomes $Y_i^0(x_i)$ or $Y_i^1(x_i)$ is observed.
We assume `no hidden
confounding' so that causal effect can be identified and confounding can be controlled for.
We learn a function $f(x,t)\rightarrow y$ that maps the context and treatment indicator to the outcome. 
%

After learning this function, we can employ an \emph{inductive} approach to estimate impact:
\begin{equation}
    im(x_i) = f(x_i, 1) - f(x_i,0).
\end{equation}
Alternatively, for data inside our observation/training set where one factual outcome is observed, impact can be \emph{transduced} via
\begin{align}
&   im(x_i) = y_i-f(x_i, 1-t_i)        \quad    \text{for} \, t_i=1,  \\
&   im(x_i) = f(x_i, 1-t_i) - y_i      \quad \text{for}\,  t_i=0.  
\end{align}

\section{Balance Regularized Neural Network Model}
\begin{figure}[h]
    \centering
    \includegraphics[width=0.95\textwidth]{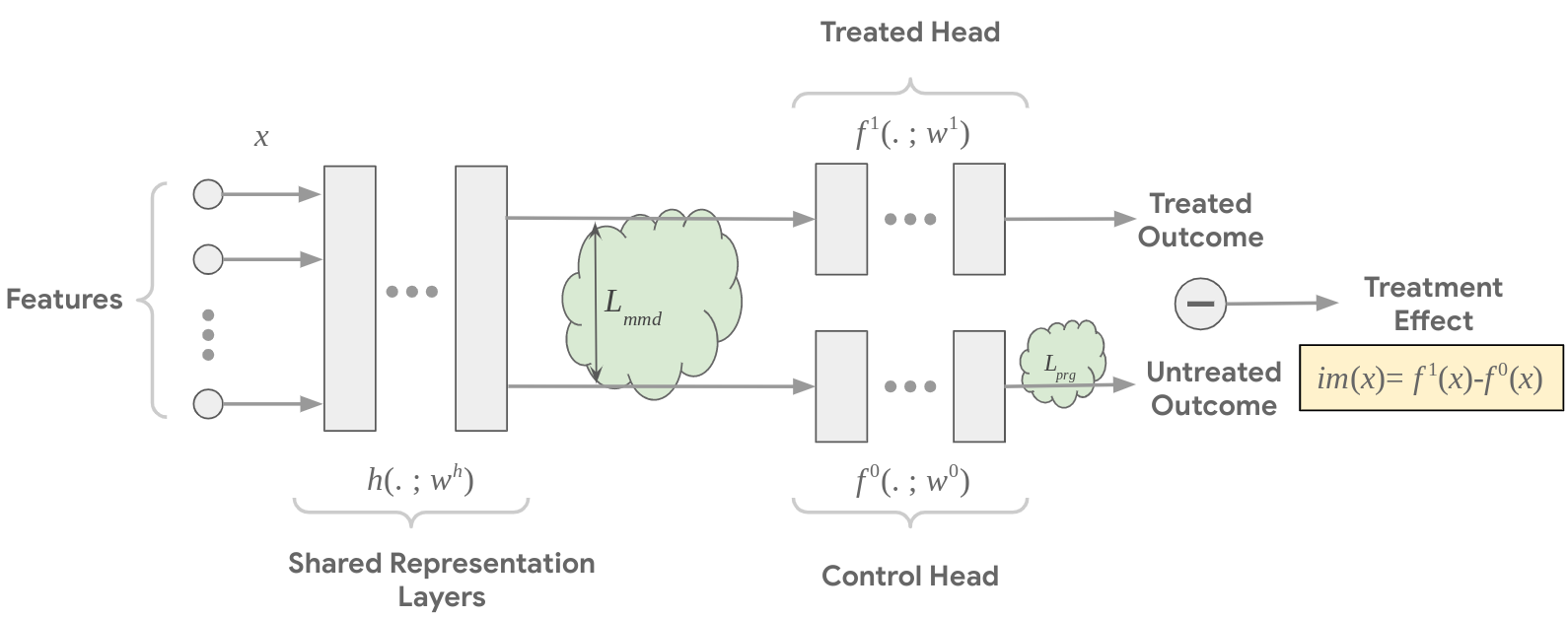}
    \caption{The architecture of our multi-head neural network model}
    \label{fig:my_label}
\end{figure}

Causal inference from observational data should take confounding factors that affect both treatment and outcome into consideration. The multi-head neural network architecture (Fig.~\ref{fig:my_label}) adjusts for confounders and prevent biased estimates inspired by the representation learning literature~\cite{bengio2013representation}. 

If the treatment assignment were random the distribution of treated units $p^1 \triangleq p\{ x | (x,t,y), t = 1\}$ would be the same as for untreated units $p^0 \triangleq p\{ x | (x,t,y), t = 0\}$. Therefore, one could train two separate models $f^0(x) \triangleq f(x, 0)$ and $f^1(x) \triangleq f(x,1)$ to predict untreated and treated outcomes respectively. However, it is common to find that in observational studies, $f^0$ and $f^1$ are trained on differing $p^0$ and $p^1$ due to confounding and selection bias, which are the most common obstacles to discovering causal relations~\cite{bareinboim2014recovering,correa2018generalized}.
Therefore, $f^0$ is not accurate for  samples drawn from $p^1$, making causal estimation, $im(x_i) = y_i - f^0(x_i)$ on triplet $(x_i, t_i=1, y_i)$ biased.
Similarly, for a data point in the control group $(x_i, t_i=0, y_i)$ the treatment model $f^1(x_i)$ may provide a confounded estimate.

The fundamental problem of causal inference  (i.e.\ only one of the two potential outcomes is observed) can be addressed under certain
assumptions~\cite{imbens2015causal,pearl2009causality} by matching each unit that received treatment ($t=1$) with its \emph{``nearest''}
units from the group that did not receive treatment ($t=0$). Propensity score matching~\cite{rosenbaum1983central} is a widely employed method for matching and grouping units based on their propensity for treatment. 
However, propensity score-based nearest neighbor matching has increasing bias with increasing dimension~\cite{abadie2006large}. To help address this, random projections~\cite{li2016matching} and local linear embedding~\cite{wang2018robust} can reduce bias by embedding the data into lower dimensional manifolds. The success of these methods inspired us using other ways of embedding and representation learning.

We learn an intermediate embedding that brings the treated and untreated distributions closer to each other before passing examples into the outcome regression part.
Inspired by recent works~\cite{johansson2016learning,shalit2017estimating,shi2019adapting} on deep causal inference, we employ a two-head shared-bottom model.

We use the function $h$ to map examples from $p^0 \cup p^1$ to a $d$-dimensional space. Because of the layer-wise structure and implicit inductive bias that stochastic gradient descent optimizer~\cite{gunasekar2018implicit,chizat2020implicit} imposes on the learning problem, the embedded individuals are restricted and the bottleneck can bring the two distributions together. This can be attributed to the middle layers that capture high level representations of concepts in the data~\cite{krizhevsky2012imagenet,le2013building} that are shared between the two groups. The subsequent heads capture distinctions between the groups that correspond to the individual outcomes.

A reasonable estimation model should be able to fit to the factual (observed) outcomes well. In other words, similar to a standard supervised learning problem, it should minimize the following objective:
\begin{equation}
   \mathcal{L}_{fit} = \frac{1}{2} \sum_{i:t_i=0} (f^0(h(x_i; w^h); w^0)-y_i)^2 + 
     \rho \frac{1}{2}  \sum_{i:t_i=1} (f^1(h(x_i; w^h); w^1)-y_i)^2, 
\end{equation}
where, $h(\cdot; w^h)$ represents the shared layers with parameters $w^h$, $f^1(\cdot; w^1)$ is the treated head with parameters $ w^1$ and $f^0(\cdot; w^0)$ is the untreated head with parameters $ w^0$. Here $\rho$ is a hyperparameter that weights either head in case the treated and untreated objectives have different size or importance. In our experiments we set $\rho$ to 1.
An explicit loss term is used to penalize the distance between distributions of the shared representation. We use Maximum Mean Discrepancy~\cite{gretton2012kernel} (MMD) as the distance measure
\begin{equation}
    \mathcal{L}_{mmd} = \sup_{||g||_\mathcal{H} \le 1} (E_{x^0 \sim p^0}[g(h(x^0))] - E_{x^1 \sim p^1}[g(h(x^1))]),
\end{equation}
where $g$ is restricted to a norm-1 ball in the Hilbert space $\mathcal{H}$ associated with a Gaussian kernel.
Further theoretical study and ablation on the choice of Kernel and MMD distance is left as an interesting future direction.

For additional balance regularization, we propose a prognostic loss to balance covariates that are predictive of the outcome by assessing the similarity in the predicted untreated outcome of both treatment and control groups. Here, the motivation (i.e.\ the prior inductive bias of the regularizer) is different from the MMD loss.
This objective helps to ensure that both groups have similar baseline outcomes without treatment. Standardized mean difference or the KS test statistic can measure differences between outcomes of groups, and differ in being parametric or nonparametric, respectively. This regularizer takes the form
\begin{equation}
    \mathcal{L}_{prg} = \text{KS}(\{f^0(h(x_i; w^h); w^0) | i: t_i=0\}, \{f^0(h(x_i; w^h); w^0) | i: t_i=1\}),
\end{equation}
where, KS is the Kolmogorov-Smirnov statistic computed on two sets of samples. 

The overall objective is then 
\begin{equation}
   \min_{w^h, w^0, w^1} \mathcal{L}_{fit} + 
  \gamma \mathcal{L}_{mmd} +
  \lambda \mathcal{L}_{prg},
\end{equation}
given the weights $\gamma$
and $\lambda$, which can take any non-negative value depending on the application. In the experiments section we will see how the choice of these weights affects bias, variance, and thus, mean squared error of the estimation. Stochastic gradient descent\footnote{ or its many variants like ADAM, ADAGRAD, RMSPROP, etc.} is used to find minimize the objective.

The proposed model architecture improves data efficiency by shared parameters across treated and untreated units. The shared parameters store commonalities between both regression tasks when optimizing for each individual task~\cite{ruder2017overview}.

The end-to-end nature of the training procedure offers an interesting approach to leverage increasing computation power, scalability, and the rise of large datasets to tackle causal questions, as evidenced by many recent works ~\cite{alaa2017deep,shalit2017estimating,louizos2017causal}.

We compare inductive and transductive approaches for estimating impact. We observed that in cases when both factual or counterfactual outcomes are available, inductive inference performs better than transductive; specifically, $f(x_i, 1)-f(x_i, 0)$ is a better estimator of ITE than $y_i-f(x_i, 0)$ or $f(x_i, 1)-y_i$. Inductive inference may reduce noise, approximation error, or bias in the shared layers. Coupling inductive inference with a multitask architecture may be an effective strategy. Moreover, this procedure can be interpreted as an empirical method for removing exogenous and unobserved effects, which may be introduced in data collection or processing, in the outcome.

\section{Experimental Results}
In this section we provide experiments to substantiate the inductive inference of impact estimation. Then, we show the promise of the proposed balance regularization framework. 

\subsection{Experimental Setup}

{\bf Datasets.} We use experimental data from the Infant Health and Development Program (IHDP), a randomized experiment that targeted low-birth-weight, premature infants, and provided the treatment group with both intensive high-quality child care and home visits
from a trained provider aiming at enhancing their IQ scores at the age of three~\cite{hill2011bayesian}. The dataset comprises 747 subjects (139 treated and 608 control), with 25 features (covariates)  associated with each instance. Outcomes are simulated based on the data generation process~\cite{hill2011bayesian}. Having simulations of both factual and counterfactual observations allows us to assess the accuracy of impact estimators. 

{\bf Measures.} The main quantity that we are interested to estimate is the individual impact $Y_i^1-Y_i^0$ and our estimate is $im(x_i)$. Recall that $Y_i^1$ and $Y_i^0$ are the potential outcomes and  $im(x_i)$ is the estimated impact.
The first measure is thus the bias of our estimate,
\begin{equation}
    Bias^2 = \mathbb{E}[(Y_i^1-Y_i^0)-im(x_i)]^2,
\end{equation}
which would be more informative along with its variance,
\begin{equation}
    Var = \mathbb{E}[(im(x_i)-\mathbb{E}[im(x_i)])^2],
\end{equation}
to express uncertainty in the estimated quantity.
The expectation is over the training data, test data, and the randomness involved in the model fitting (e.g. initialization and SGD noise). These two quantities are combined to form a Mean Squared Error (MSE) of the estimate:
\begin{equation}
    MSE = Bias^2 + Var.
\end{equation}
In the causal inference literature the MSE is sometimes replaced by  Precision in Estimation of Heterogeneous Effect (PEHE) which is defined as
\begin{equation}
   PEHE = \sqrt{\frac{1}{n}\sum_{i=1}^n (im(x_i) - (Y_i^1-Y_i^0))^2}.
\end{equation}
It is easy to verify that $\sqrt{MSE} = PEHE$.

{\bf Neural network structure and training.} For all models we used Multilayer Perceptron  (MLP) architecture.
The two head model contains a shared bottom with two hidden layers of 20 neurons and two heads, which each have a 20-neuron hidden layer and a 1-neuron output. 
For the separate head model we used two MLP networks, one for fitting the treatment group and one for fitting the control group. They include three hidden layers each with 20 neurons with a one neuron output layer. The input layer has 25 neurons in both cases. All  layers except the final layer use ReLU nonlinearities. None of batch norm, dropout, weight decay, data augmentation, and normalization are employed. 
We used ADAM~\cite{kingma2014adam} with a learning rate of $0.0001$, a batch size of 64, and $50$ epochs for training. Out of $\approx 750$ examples, $20 \%$ are randomly held for the test set. To simulate different training sets we randomly selected $60 \%$ of data 5 times and for each we run 4 times with random initialization. Therefore, our results are averaged over 20 random runs with $\approx 450$ and $\approx  150$ training and test examples, respectively.

\subsection{Results}

Table~\ref{tb:all_results} shows the average squared bias, average variance, and the mean and standard deviation of the MSE error over 20 iterations of a simulated set of potential outcomes over the IHDP dataset. No regularization is applied for this table.
First, we note that incorporating the shared bottom in the two head model improves mean and standard deviation of the MSE.
Second, the two head model improves MSE with a significant reduction in the variance and a relatively small increase in bias. 
Third, inductive inference consistently leads to more accurate impact estimation in both of separate and two-head models compared with transductive inference.

This lower bias of separate head model (in inductive setting) can be explained by the fact that two separate neural networks are more flexible to fit to data compared to the case where their bottom is restricted to be the same. This flexibly comes with the cost of significantly higher variance.

\begin{table}[h]
\vspace{-2mm}
\centering
\caption{Numerical comparison of the two head model with separate head model}
\label{tb:all_results}
\begin{tabular}{ccccc}
    \toprule
	 Model & Inference Mode &  $Bias^2$ & $Var$ & $MSE$ (average $\pm$ std)\\
	 	 	\midrule
	 \multirow{2}{*}{separate head} & transductive & $6.15$ &	$0.72$ &	$6.87 \pm	0.89$  \\
	 	  & inductive & $2.87$ &	$1.00$ &	$3.87 \pm	0.85$ \\
	\midrule
	 \multirow{2}{*}{two head}  & transductive & $4.55$ &	$0.23$ & $4.78 \pm 0.19 $ \\
	   & inductive & $3.34$	& $0.32$ & $3.66 \pm 0.23 $ \\
	\bottomrule
\end{tabular}
\end{table}

\begin{figure}[h]
  \centering
\begin{subfigure}{.4\textwidth}
  \centering
  \includegraphics[width=1\textwidth]{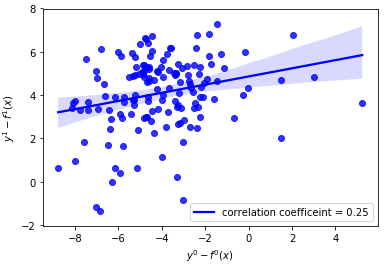}
  \caption{Separate head model}
  \label{fig:scatter-sep}
\end{subfigure}
\begin{subfigure}{.4\textwidth}
  \centering
  \includegraphics[width=1\textwidth]{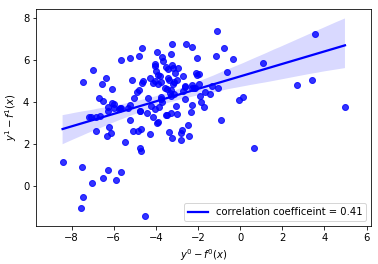}
  \caption{Two head model}
  \label{fig:scatter:two}
\end{subfigure}
    \caption{The correlation between real and prediction values of control and test group}
    \label{fig:scatter}
\end{figure}

Inductive inference forgoes variance within the observed outcome  and relies on predictions of both outcomes to estimate impact. To further study the reasons behind its superiority we have plotted the difference between real and predicted treated outcome versus the difference between real and predicted control outcome in Fig.~\ref{fig:scatter}. The positive correlation shows that whenever the predicted treated value ($f^1(x)$) is larger than the real treated value ($y^1$), then the predicted control ($f^0(x)$) is larger than the observed control value ($y^0)$. Both prediction heads have a bias that is positively correlated and may be canceled out by the inductive difference $f^1(x)-f^0(x)$. Further theoretical study of this phenomenon is left for future work.
Another expected, yet interesting, observation in this figure is that the predictions on the treated values underestimate the observed treated value while the predictions on the control value overestimate the observed control values. Both models attempt to statistically fit to the data by pushing the predictions towards their target values, and he optimum model converges somewhere in between. 

\begin{figure*}[h] 
\centering
\begin{subfigure}{.3\textwidth}
  \centering
  \includegraphics[width=1\linewidth]{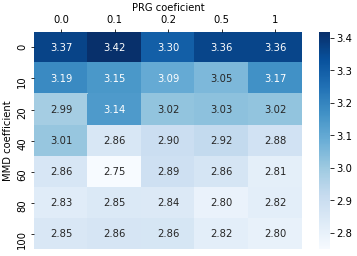}
  \caption{$Bias^2$}
  \label{fig:reg-bias}
\end{subfigure}%
\begin{subfigure}{.3\textwidth}
  \centering
  \includegraphics[width=1\linewidth]{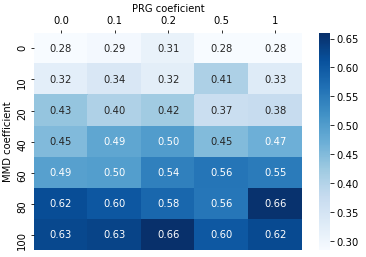}
  \caption{$Var$}
  \label{fig:reg-var}
\end{subfigure}
\begin{subfigure}{.38\textwidth}
  \centering
  \includegraphics[width=1\linewidth]{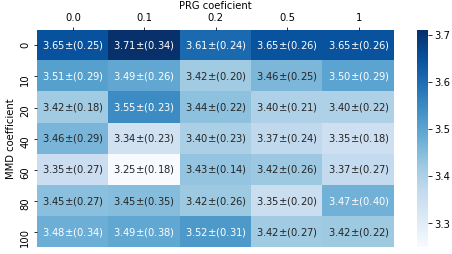}
  \caption{$MSE$ (average $\pm$ std)}
  \label{fig:reg-mse}
\end{subfigure}
\caption{The bias, variance and MSE of the two head model with respect to varying the regularization weights. The lowest MSE occurs at the suitable trade-off between the bias and variance.}
\label{fig:regularizers}
\end{figure*}

Next, we show the effectiveness of balance regularization by increasing the coefficient of MMD and PRG losses from 0 to larger values in the two head model. The estimation error in terms of bias, variance and MSE are shown in
Fig.~\ref{fig:regularizers}. 
We can see the bias is generally decreasing from top to bottom (increasing the MMD regularization) and from left to right (increasing prognostic regularization) in Fig.~\ref{fig:reg-bias}.
Conversely, the variance of the predictions demonstrated in Fig~\ref{fig:reg-var} increases when enforcing more balance regularization. This creates an interesting trade-off between bias and variance that potentially leads to an optimum MSE in the middle of the coefficients grid as shown in Fig~\ref{fig:reg-mse}.
It is notable that the different ranges for MMD and PRG coefficient are due to difference in the magnitude of the associated loss terms.
Fig.~\ref{fig:regularizers} shows that performance is more sensitive to MMD regularization than PRG.



\section{Related work}
Machine learning techniques have already been used in estimating causal effects from data~\citep{athey2015machine,jung2020learning, malinsky2017estimating,linden2016combining,peters2017elements}.
In addition to aforementioned works work~\cite{johansson2016learning,shi2019adapting}, the idea of deep representation learning has been embraced by many recent studies such as~\cite{li2017matching,yao2018representation,schwab2018perfect,chen2019deep,schwab2020learning}. These ideas have been extended to leverage propensity score~\cite{alaa2017deep,shi2019adapting} or to work with instrumental variables~\citep{hartford2017deep}. Machine learning is contributing to causal effect estimation via many recently developed deep learning techniques such as Generative Adversarial Networks (GANs)~\citep{yoon2018ganite}, Knowledge Distillation~\citep{makar2019distillation}, Adversarial Training~\citep{du2019adversarial}, Meta Learning~\citep{sharma2019metaci}, and Transfer Learning~\citep{kunzel2018transfer}.

The problem of estimating causal effects is about learning the effect of an intervention, the treatment.
Interventions modify the distribution, and here we have two, namely the treated and the untreated distributions.  
The machine learning field has investigated related settings.  Learning with distribution shift~\citep{quionero2009dataset} considers it generally, and categorizes the distribution shift into different types, such as a shift in the input distribution (covariate shift), or a shift in the output distribution (target shift), as well as other types.   Many of these shifts can occur simultaneously in the context of causal effect estimation, and result in confounding bias.

More specific settings include those of domain invariance, domain transfer, and domain adaptation~\citep{zhang2015multi}. 
They involve learning across multiple domains, corresponding to different distributions.  For example, they consider how to transfer learning from the distribution of one domain to a different one~\citep{magliacane2018domain, zhang2020domain}.  Learning domain invariance is perhaps the most closely related to ours: invariant risk minimization~\citep{arjovsky19invariant} attempts to learn a representation that is invariant across domains.  This has also been formulated adversarially, where representations are penalized if they differ across domains~\citep{ganin2016domain}.  Representations can also be made similar across domains via GANs~\citep{yoon2018ganite}.
The area of causal discovery from shifted distributions~\citep{huang2020causal} can also lead to interesting insights about causal effect estimation.

Our setting differs from the above in that we know that the source of the distribution shift is due to the treatment, in particular the treatment assignment, and its effect on outcome.  Also, we are focused on predicting the difference in outcomes for different treatments, rather than absolute outcomes. This is what allows pinpointing the distribution shift using techniques from statistics, such as the propensity score and prognostic score~\citep{hansen08prognostic}.  Propensity score has been used as an input feature for causal inference scenarios~\cite{hill2011bayesian}, and also as an auxiliary neural network head~\cite{shi2019adapting}.  Prognostic score has been used in epidemiology for evaluating the quality of propensity score matching~\citep{stuart2013prognostic}, but has not previously been used for regularization, or with neural networks.

\section{Conclusion and Future Work}
We have proposed a deep neural network model for estimating treatment impact in observational studies. Our balance regularization is inspired by matching the treatment and control distributions and is built on the top of representation learning and enjoys the end-to-end training and sample efficiency of gradient based neural network models while being sufficiently flexible to model complex functions.
We acknowledge that our empirical experiments have been conducted on a relatively small sample size. Extending this work to larger datasets and systematically studying the effects of balance regularizers combined with other regularizers (like dropout and weight decay) is left as future work. 
One can potentially make the prognostic loss more symmetric by enforcing it on the treatment head too. Another interesting line of work is combining these losses with treatment assignment prediction on as a third head~\citep{shi2019adapting}.


\bibliographystyle{unsrt}
\bibliography{references.bib}

\end{document}